\documentclass[sigconf]{acmart}

\usepackage{xspace}

\makeatletter
\DeclareRobustCommand\onedot{\futurelet\@let@token\@onedot}
\def\@onedot{\ifx\@let@token.\else.\null\fi\xspace}

\def\eg{\emph{e.g}\onedot} 
\def\ie{\emph{i.e}\onedot}

\def\wrt{w.r.t\onedot} 
\def\etal{\emph{et al}\onedot}
\makeatother

\AtBeginDocument{%
  \providecommand\BibTeX{{%
    \normalfont B\kern-0.5em{\scshape i\kern-0.25em b}\kern-0.8em\TeX}}}

\setcopyright{acmcopyright}
\copyrightyear{2018}
\acmYear{2018}
\acmDOI{10.1145/1122445.1122456}

\acmConference[ACM Multimedia Conference 2020]{ACM Multimedia Conference 2020: the 28th ACM International Conference on Multimedia}{October 12-16, 2020}{Seattle, United States}
\acmBooktitle{ACM Multimedia Conference 2020: the 28th ACM International Conference on Multimedia,
October 12-16, 2020, Seattle, United States}
\acmPrice{15.00}
\acmISBN{978-1-4503-XXXX-X/18/06}

\begin{document}

\title{Memory-Augmented Relation Network for Few-Shot Learning}





\author{
   Jun He$^{1}$, 
   Richang Hong$^{1}$, 
   Xueliang Liu$^{1}$, 
   Mingliang Xu$^{2}$, 
   Zhengjun Zha$^{3}$, 
   Meng Wang$^{1}$
}
\affiliation{
   \institution{$^{1}$Hefei University of Technology, Hefei, China}
   \institution{$^{2}$Zhengzhou University, Zhengzhou, China}
   \institution{$^{3}$University of Science and Technology of China, Hefei, China}
}

\renewcommand{\shortauthors}{He and Hong, et al.}

\begin{abstract}
   Metric-based few-shot learning methods concentrate on learning transferable feature embedding that generalizes well from seen categories to unseen categories under the supervision of limited number of labelled instances. However, most of them treat each individual instance in the working context separately without considering its relationships with the others. In this work, we investigate a new metric-learning method, Memory-Augmented Relation Network (MRN), to explicitly exploit these relationships. In particular, for an instance, we choose the samples that are visually similar from the working context, and perform weighted information propagation to attentively aggregate helpful information from the chosen ones to enhance its representation. In MRN, we also formulate the distance metric as a learnable relation module which learns to compare for similarity measurement, and augment the working context with memory slots, both contributing to its generality. We empirically demonstrate that MRN yields significant improvement over its ancestor and achieves competitive or even better performance when compared with other few-shot learning approaches on the two major benchmark datasets, \ie \textit{mini}Imagenet and \textit{tiered}Imagenet.
\end{abstract}




\begin{teaserfigure}
  \includegraphics[width=\textwidth]{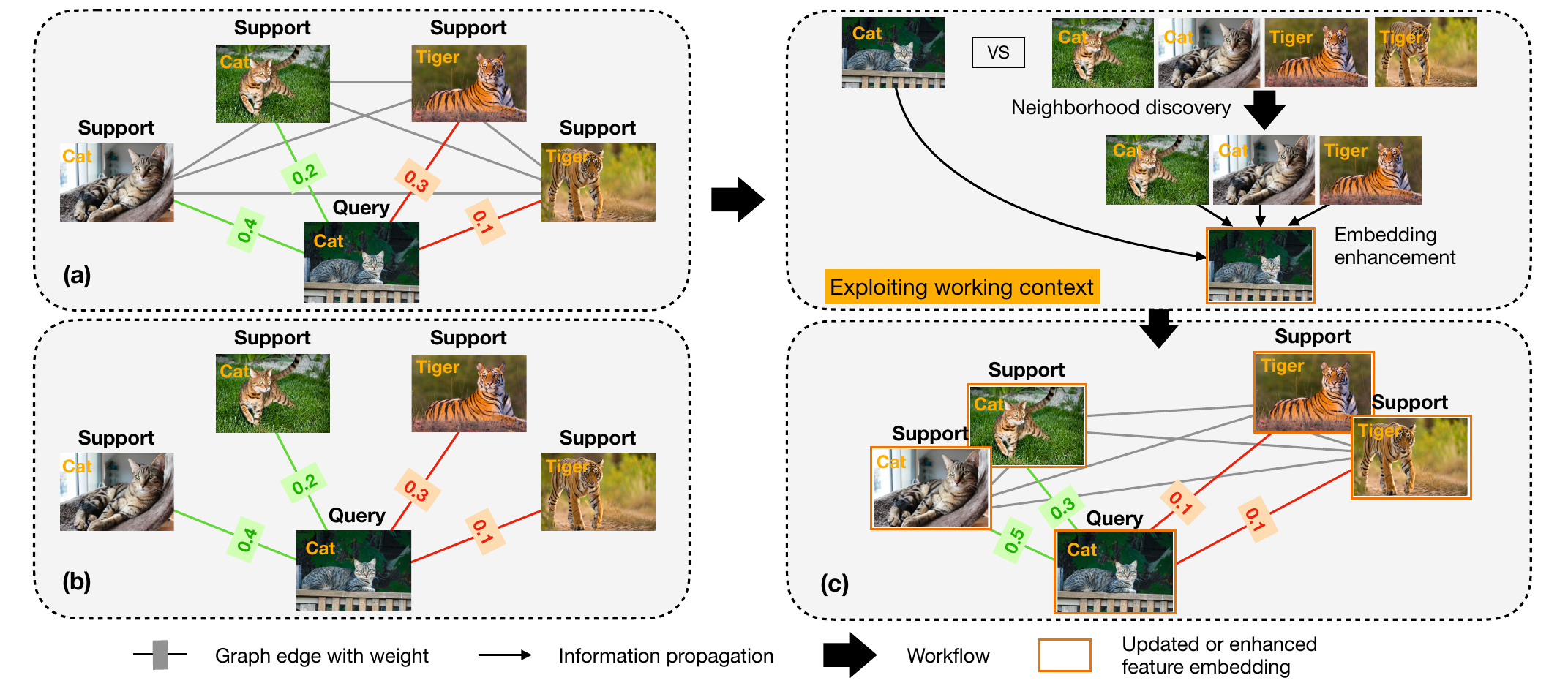}
  \vspace{-1.5em}
  \caption{Major motivation. Conventional metric learning methods for few-shot learning focus on the query-support relations in (b) to recognize a query sample but ignore the support-support relations in (a). We notice that explicitly exploiting these relations results in more discriminative features, as shown in (c), which in turn improves performance.}
  \Description{We aim to estimate sample-sample relationships and exploit the working context for embedding enhancement.}
  \label{fig:teaser}
\end{teaserfigure}

\maketitle

\section{Introduction}
Deep artificial agents have achieved impressive performance on various computer vision tasks like classification~\cite{szegedy2016rethinking,he2016deep,huang2017densely}, object detection~\cite{he2017mask,ren2015faster,lin2017feature} and semantic segmentation~\cite{chen2018encoder,zhang2018context,huang2019ccnet}. For classification, deep recognizers have outperformed humans on visual recognition challenge for years~\cite{he2015delving}. However, the success hinges on the ability to apply gradient-based optimization routines to high-capacity models when given large-scale training samples as supervision~\cite{santoro2016meta}. When the training samples are scarce, it becomes challenging to learn to recognize new concepts due to the overfitting problem. In contrast, we human can learn new concepts fast with only few examples. The gap here presents the few-shot visual recognition task.

Few-shot visual recognition, also termed few-shot learning, aims to learn novel visual concepts from one or a few labelled instances. It was first introduced by Fei-Fei \etal~\cite{fe2003bayesian} in 2003, and has attracted lots of attention ever since. Data augmentation is a straightforward method to tackle the task~\cite{hariharan2017low,zhang2019few}, but it does not solve the low-data problem. A promising trend is to transfer knowledge from known categories (\ie seen categories with abundant training examples) to new categories (\ie novel categories with few examples)~\cite{fe2003bayesian,rodner2010one,yu2010attribute}, simulating the human learning process~\cite{gentner1997reasoning, zhou2019learning}. To this end, a variety of approaches have been proposed to learn transferable knowledge in various forms, among which metric-learning methods that concentrate on learning transferable feature embedding fall into one of the main strands.

Given an unlabeled query sample and a limited number of labelled support samples, a metric-learning method first maps all the input samples into a latent space in which similarities between the query sample and each support sample are estimated according to a predefined distance metric, \eg cosine distance metric~\cite{vinyals2016matching} or Euclidean distance metric~\cite{snell2017prototypical}. It then labels the query sample as same as the support sample that gets the highest similarity score~\cite{vinyals2016matching, qi2018mmnet, li2019memory}. While promissing, most metric-learning methods merely consider instant pairwise query-support relationships as shown in Figure~\ref{fig:teaser}(b) but fail to explore support-support relationships among the labelled support samples, let alone that among the unlabeled ones. Moreover, the \textit{single-image-estimated} \footnote{The image representation is independently estimated based on a single image} features they used for few-shot learning are normally not discriminative enough to bring a considerable performance improvement. We also doubt the fact that the selected metric suits few-shot learning best. A predefined metric may tailor feature extractor to generate a latent space that matches this very metric perfectly but is sub-optimal for few-shot learning. 

To address the above problems, we extend~\cite{sung2018learning} to explicitly explore the relationships between each two samples and propose to propagate information from relevant samples for embedding enhancement. Particularly, we adopt a parameterized relation module to estimate the relationship between two samples. The module takes two samples as input and tells to what extent they belong to the same category. It is trained to produce close relationships for similar samples and distant relationships for dissimilar samples. With this module, we aim to learn a generic distance metric simutaneously when searching for the optimal latent space for few-shot learning. A weighted fully connected relation graph as illustrated in Figure~\ref{fig:teaser}(a) then is constructed by applying the learnt metric for pairwise comparison in the working context. We propose to explore this relation graph for embedding enhancement to facilitate few-shot learning. For enhancing the representation of an instance in the graph, we retrieve its neighbors and perform weighted information propagation to attentively aggregate visual information from the neighbors. Such an information propagation reduces noise in class representations and expands decision boundaries, actually serving as a manifold smoothing regularization. Additionally, we add a memory to serve as the working context which controls the number of samples that are available in the graph construction and information propagation. Because the memory has a flexible memory capacity, it can easily extend to hold unlabeled query samples or unlabeled auxiliary samples, making it generalize well to transductive setting or semi-supervised setting. 

We dub the proposed method Memory-Augmented Relation Network (MRN). Our main contributions are as follows:
\begin{itemize}
   \item A generic distance metric is learned for few-shot learning. By designing the distance metric as a parameterized relation module, we learn transferable representations and distance metric end to end.
   \item We propose to enhance representations via attentively aggregating information from the neighborhood context. The enhanced representations are shown to be more discriminative. 
   \item We propose a general memory-based framework for few-shot learning. It comes with tightly coupled structure in which the propagation procedure and the classifier share the same learnable distance metric.
   \item We evaluate the proposed MRN on two benchmark dataset, namely \textit{mini}Imagenet~\cite{snell2017prototypical} and \textit{tiered}Imagenet~\cite{ren2018meta}. It achieves great improvements over existing methods in few-shot setting.
\end{itemize}


\section{Related Work}
Approaches that aim to tackle few-shot learning can be roughly divided into three groups: meta-learning approaches, metric-learning approaches, and hallucination based approaches. Recently, transductive learning has drawn much attention because of the performance boost. In this section, we briefly review relevant work for each of the groups and point out their relevance to our method.

\noindent\textbf{Meta-learning approaches} Meta-learning approaches usually utilize optimization-based meta-learning or the learning-to-learn paradigm~\cite{hochreiter2001learning,andrychowicz2016learning} to train a meta-learner for few-shot learning. The meta-learner either learns an optimizer for training another linear classifier~\cite{ravi2016optimization} or learns an optimal initialization state for the base classifier~\cite{finn2017model,andrei2019leo,kwonjoonlee2019metaoptnet}, from which fast learning is feasible with few examples. The meta-learner is usually difficult to train due to its complicated memory architectures (\eg RNNs, recurrent neural networks) and the temporally-linear hidden state dependency~\cite{mishra2018simple}. Besides, an additional fine-tuning on the target few-shot learning task is further required. In contrast to those methods, the proposed MRN can be easily trained end to end from scratch without bothering to introduce any complicated memory architecture, and can adapt to new tasks directly with no need for fine-tuning.

\noindent\textbf{Metric-learning approaches} Metric-learning approaches mainly aim at learning transferable representations for few-shot learning~\cite{vinyals2016matching,snell2017prototypical,sung2018learning, koch2015siamese, ziyang2019parn,li2019revisiting,qi2018mmnet}. To this end, Vinyals \etal~\cite{vinyals2016matching} presented matching networks which use an attention mechanism to derive a weighted k-NN classifier, and devised an episodic meta-learning mechanism to train the classifier for fast adaptation. Snell \etal~\cite{snell2017prototypical} proposed to take the mean representation of support samples in each class as prototype and recognize a query sample according to its square euclidean distances against the prototypes. Li~\etal~\cite{li2019revisiting} advanced this image-to-category measure by replacing image-level representation with a bunch of local descriptors. To overcome local connectivity, Wu~\etal~\cite{ziyang2019parn} also exploited local descriptors to deal with inherent local connectivity with a dual correlation attention mechanism. They proposed to concatenate four globally related features derived from cross-correlation attention and self-correlation attention to yield a more representative feature. Different from~\cite{ziyang2019parn} that focuses on digging a more representative representation out of one single image, in this work, we propose to enhance the representation of a sample via aggregating information from its neighborhood, \ie the other samples. Both~\cite{ziyang2019parn} and our MRN are descendants of relation network~\cite{sung2018learning}, a deep model which extends siamese network~\cite{koch2015siamese} by adopting the innovative episodic meta-training mechanism proposed in~\cite{vinyals2016matching}.

\noindent\textbf{Hallucination based approaches} For a recognizer that has a feature extractor and a classifier as two distinctive components, hallucination based approaches propose to improve performance from two aspects: data augmentation~\cite{hariharan2017low,wang2018low,zhang2019few} and classifier weight vector hallucination~\cite{qi2018low,qiao2018few,zhou2019learning,gidaris2018dynamic,gidaris2019generating}. The data augmentation approaches commonly focus on expanding intra-class variations by hallucinating new training samples. The classifier weight vector hallucination approaches, on the other hand, propose to directly set weight vector for a novel category in the classifier. Such approached are inspired by the close relationships between classifier weight vectors and the feature representations associated with the same category produced by the feature extractor. Though straightforward, the imprinting process surprisingly provides an instant good classification performance on novel categories and an initialization for future fine-tuning. 

\noindent\textbf{Transductive learning} In few-shot learning, the recognition of each query sample is carried out independently one by one with only a few labelled support samples. Transductive few-shot learning proposes to feed all the query samples once at a time and predict them as a whole. In this setting, the relationships among all samples, both labelled and unlabeled ones, can be considered to improve performance. For example, Liu \etal~\cite{liu2018learning} presented a model that utilizes the entire query set for transductive inference, which propagates labels from labelled samples to unlabeled ones based on visual affinities. Other representatives are~\cite{ye2018learning, li2019memory,kim2019edge}. In~\cite{ye2018learning}, Ye~\etal trained an attention-based transformer to transform task-agnostic embeddings to task-specific embeddings for few-shot learning. The proposed FEAT selects relevant instances and combines their transformed embeddings to obtain a new task-specific embedding. Li~\etal~\cite{li2019memory} followed the feature propagation idea and proposed to aggregate information from neighborhood in a tree graph. Likewise, we also propose to aggregate information from the neighborhood context for representation enhancement. The key difference between~\cite{li2019memory} and our MRN is that we extend~\cite{sung2018learning} to jointly learn a generic distance metric and update representations with information from samples selected by this very distance metric. 

\begin{figure*}[t]
   \begin{center}
      \includegraphics[width=\linewidth]{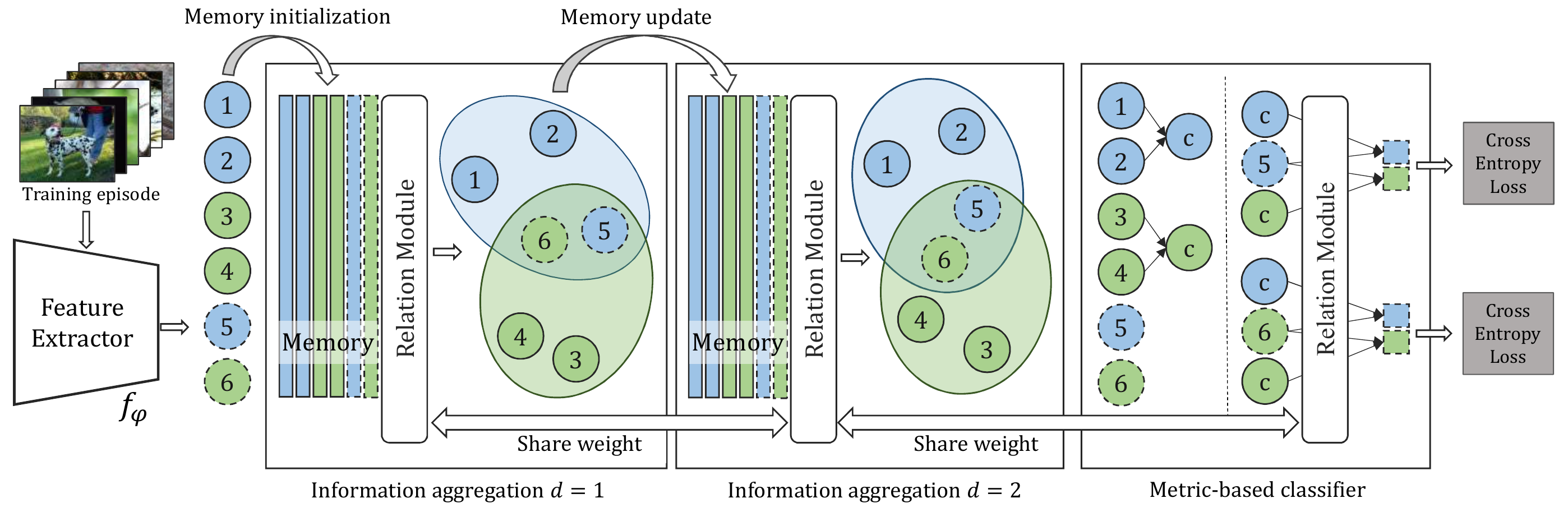}
   \end{center}
   \caption{The overall framework of the proposed MRN model. In this illustration, we present a 2-shot 2-way problem as an example. Circles with solid line denote nodes corresponding to support images, and the ones with dashed line are the query images to be classified. Two different colors represent two categories. For brevity, we leave the edges in graph not drawn. The distance between two nodes reflects their relationship. The detailed process is described in Section~\ref{sec:method_overview}. Best viewed in color.}
   \label{fig:framework}
\end{figure*}

\section{Preliminary}
\label{sec:fsl-definition}
A few-shot learning task $\mathcal{T}$ is generally composed of two sets of instances, the support set $\mathcal{S}={\{(x_i^s, y_i^s)\}}_{i=1}^{N_{\mathcal{S}}}$ that contains $N_{\mathcal{S}}$ labelled instances and the query set $\mathcal{Q}={\{x_i^q\}}_{i=1}^{N_{\mathcal{Q}}}$ which contains $N_{\mathcal{Q}}$ unlabeled instances. For simplicity, we consider the well-organized $K$-shot $C$-way few-shot setting, where the support set $\mathcal{S}$ is prepared by sampling $K$ instances from each of the $C$ categories. Instances in query set $\mathcal{Q}$ are sampled from these $C$ categories as well but must share no individual instance with $\mathcal{S}$, \ie $\mathcal{S}\cap\mathcal{Q}=\emptyset$. Few-shot learning aims to classify each unlabeled instance in $\mathcal{Q}$ with labelled instances in $\mathcal{S}$ as supervision. 

Intuitively, one can train a classifier on the support set in hope that it generalizes well to the query set. Unfortunately, this straightforward proposal suffers from severe overfitting due to the data scarcity problem. The community usually resorts to an auxiliary meta-train set $\mathcal{D}_{\textit{base}}$ to learn transferable knowledge to improve generalization on $\mathcal{Q}$. The set $\mathcal{D}_{\textit{base}}$ contains abundant labelled examples from $C_{\textit{base}}$ categories and has a disjoint label space \wrt the target few-shot learning task, \ie $C_{\textit{base}} \cap C = \emptyset$. An effective way to exploit this meta-train set is to perform episodic meta-training with a large amount of sampled training episodes, as proposed in \cite{vinyals2016matching}. 

To set up a training episode $\hat{\mathcal{T}}$, we first randomly sample $C$ categories from $C_{\textit{base}}$. Then, for each of the $C$ categories, we randomly sample $K$ labelled examples to serve as the support set $\hat{\mathcal{S}}$. A fraction of the remainder examples of the $C$ categories are selected to act as the query set $\hat{\mathcal{Q}}$. Obviously, the training episode is actually mimicking the target $K$-shot $C$-way few-shot learning task $\mathcal{T}$. Training a model over tens of thousands of such training episodes with the objective defined in Equation (\ref{eq:training_objective}) would yield one model that generalizes well when applied to tasks that contain novel categories.
\begin{equation}
   \resizebox{0.9\linewidth}{!}{
      $
      \hat{\theta} = \arg\max\limits_{\theta} E_{\hat{\mathcal{T}}}
         \begin{bmatrix}
            \frac{1}{\vert\hat{\mathcal{Q}}\vert}
            \sum\limits_{(x,y)\in\hat{\mathcal{Q}}} \log P_{\theta}(y|x, \hat{\mathcal{S}})
         \end{bmatrix} + \mathcal{R}(\theta)
      $
   }
\label{eq:training_objective}
\end{equation}
In Equation (\ref{eq:training_objective}), $\theta$ are the parameters of the model, $P_{\theta}(y|x, \hat{\mathcal{S}})$ denotes the probability of sample $x$ being from category $y$, and $\mathcal{R}(\theta)$ is the standard regularization. Following \cite{vinyals2016matching,snell2017prototypical,sung2018learning,liu2018learning}, we adopt the episodic meta-training strategy in this work.

\section{Method}
\label{sec:method_overview}

We present a memory-augmented relation network (MRN) that has the pipeline as illuastrated in Figure~\ref{fig:framework} to solve few-shot learning problem. It basically consists of four components: a feature embedding function $f_{\varphi}: \mathbb{R}^{W \times H \times 3} \rightarrow \mathbb{R}^d$ for mapping similar images to nearby points in latent space, an information aggregation component together with an episodic memory for representation enhancement, a relation module $g_{\phi}: \mathbb{R}^d \rightarrow \mathbb{R}$ that learns to tell whether two images are from the same category, and a metric-based classifier for making the final predictions. In the following sections, we explain the information aggregation component and the metric-based classifier in detail and give a brief description of the relation module. The embedding function $f_{\varphi}$ is left untouched because MRN is capable of working with all possible backbones.

\subsection{The Metric-based Classifier}
\label{sec:metric_based_classifier}
We first introduce our parameter-free metric-based classifier in order to make a clear impression. The classifier, in one word, takes an unlabeled query image and a bunch of labelled support images as input, and goes through an image-to-category measure to assign the query image to the category that it most likely belongs to. Let $\mathbf{f}=\{\mathbf{f}_1,\dots,\mathbf{f}_{\vert\hat{\mathcal{T}}\vert}\}$ be the feature embeddings of instances in a given $K$-shot $C$-way training episode $\hat{\mathcal{T}}$, it imposes the cross entropy loss on each image for network training:
\begin{equation}
   \mathcal{L}_i = - \sum\limits_t^C y_{i,t} \log(\frac{\exp(- D_{\mathcal{C}}(\mathbf{f}_i,\mathbf{c}_t))}{\sum_j^C \exp(- D_{\mathcal{C}}(\mathbf{f}_i,\mathbf{c}_j))})
   \label{eq:cross_entropy_train}
\end{equation}
where $\mathbf{c}_t$ represents the $t$-th class centroid that is computed as the mean value of all support images in the category as defined in~\cite{snell2017prototypical}, $y_{i,t}$ is the index label with $y_{i,t}=1$ if the image $\mathbf{f}_i$ belongs to the $t$-th category and $y_{i,t}=0$ otherwise, and $D_{\mathcal{C}}$ denotes a certain distance metric. Likewise, let $\mathbf{f}=\{\mathbf{f}_1,\dots,\mathbf{f}_{\vert{\mathcal{T}}\vert}\}$ be the feature embeddings of instances in the target $K$-shot $C$-way few-shot learning task ${\mathcal{T}}$, in the test phase, it labels each query image according to its distance to the class centroids as defined in Equation~\ref{eq:classifier}.
\begin{equation}
   \hat{y}_i = \arg \min\limits_{t} D_{\mathcal{C}}(\mathbf{f}_i, \mathbf{c}_t), ~~t = 1,2,\dots, C
   \label{eq:classifier}
\end{equation}

\subsection{Embedding Enhancement within Episodic Memory}
\label{sec:embedding_enhancement_within_memory}
The metric-based classifier highly relies on discriminative representations to predict correct labels. However, in few-shot learning, the embedding function is usually not sufficiently trained to model representative semantics in the training data. When feeding these raw features into a metric-based classifier, it leads to inferior performance. To alleviate the problem, we propose to enhance a sample by adding extra information propagated from other similar or relevant samples to its representation. Specifically, when working with episodic meta-training, for each image $x_i$ in a training episode $\hat{\mathcal{T}}=\hat{\mathcal{S}}\cup\hat{\mathcal{Q}}$, we select its $k$ nearest neighbors $\mathcal{A}_i\in\mathbb{R}^{k \times d}$ from available samples $\mathcal{M}\in\mathbb{R}^{m \times d}$ and update its feature embedding via gathering information from these neighbors as in Equation~\ref{eq:weighted_aggregation} that followed. 
\begin{equation}
   \mathbf{f}_i = \lambda\mathbf{f}_i + (1-\lambda)\sum_{\mathbf{f}_j\in\mathcal{A}_i} w_{i,j}\mathbf{f}_j, ~~\mathbf{f}_i\notin\mathcal{A}_i
   \label{eq:weighted_aggregation}
\end{equation}
\begin{equation}
   \begin{aligned}
   w_{i,j} = \frac{\exp(-D_{\mathcal{G}}(\mathbf{f}_i, \mathbf{f}_j))}{\sum_{\mathbf{f}_k\in\mathcal{A}_i}\exp(-D_{\mathcal{G}}(\mathbf{f}_i, \mathbf{f}_k))}, ~~\text{iff}~ \mathbf{f}_j,\mathbf{f}_k\in\mathcal{A}_i
   \end{aligned}
   \label{eq:weight_coefficient}
\end{equation} 
where $\mathbf{f}_i$ denotes the representation of the instance $x_i$, $w_{i,j}$ is the coefficient that controls information flow during the aggregation as defined in Equation~\ref{eq:weight_coefficient},  $D_{\mathcal{G}}$ represents some distance metric, and the hyperparameter $\lambda$ indicates the proportion of information preserved to avoid disturbance from neighborhood.

The information propagation could be naturally simulated by message passing in graph. We therefore organize all instances $\{\mathbf{f}_{\varphi}(x_i)\} \cup \mathcal{M}$ as a weighted relation graph $G=(V,E)$ as defined in Equation~\ref{eq:graph_construction}:
\begin{equation}
\mathbf{v}_i = \mathbf{f}_i, ~~ e_{i,j} = D_{\mathcal{G}}(\mathbf{f}_i, \mathbf{f}_j) ~~~ \forall \mathbf{f}_i,\mathbf{f}_j\in\{\mathbf{f}_{\varphi}(x_i)\} \cup \mathcal{M}
\label{eq:graph_construction}
\end{equation}
where the edge $e_{i,j}$ reflects the visual affinity between sample $x_i$ and $x_j$. For each node $\mathbf{v}_i \in G$, we select its $k$-nearest neighbors based on edge weights and perform the aforementioned weighted information aggregation to update its node embedding, which accordingly updates image representation $\mathbf{f}_i$ because the two are identical. Theoretically, the information aggregation can be performed iteratively:
\begin{equation}
   \mathbf{v}_i^{d} = \text{AGGREGATE}(\mathcal{A}_i^{d-1})
   \label{eq:iterative_aggregation}
\end{equation}
where $d$ denotes the current search depth. By increasing $d$, we can aggregate information from more samples, like the neighbors of neighbors. Consequently, the deeper we go into the graph, the denser the updated embeddings are. However, a big $d$ hurts the classification performance. Imagining the situation where $d$ is big enough, all images collapse to the same point in the latent space which makes the classification impossible. 

Undoubtedly, the number of potentially exploitable instances in $\mathcal{M}$ matters in the propagation procedure. In this work, we keep an episodic memory to serve as $\mathcal{M}$ and consider a more controllable propagation procedure for embedding enhancement. The memory has an adjustable memory size in case it can hold flexible number instances. As depicted in Figure~\ref{fig:framework}, we initialize the memory with features extracted by the embedding function $f_{\varphi}$, and update the memory whenever the information aggregation changes the feature stored in one memory slot. With the help of the memory cache, our embedding enhancement adapts very easily to transductive few-shot learning or semi-supervised few-shot learning by initializing the memory to hold unlabeled samples that we intend to exploit during the weighted information propagation. Different from other work that adopt memory~\cite{santoro2016meta,qi2018mmnet}, the memory in our work lives as long as an episode exists but cannot survive across episodes. When a new episode arrives, the old memory is destroyed and a new memory initialized with new data is created.

\subsection{Relation Module: Learning to Compare}
\label{sec:relation_module}
\begin{figure*}
   \begin{center}
      \includegraphics[width=\linewidth]{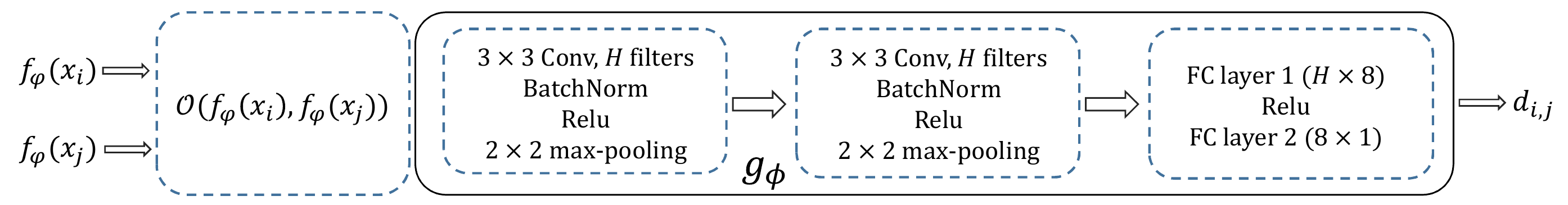}
   \end{center}
   \vspace{-0.8em}
   \caption{ Architecture of our relation module. It consists of two convolutional blocks and two fully connected layers.}
   \vspace{-0.5em}
   \label{fig:relation_module}
\end{figure*}
We design the relation module as simple as the parameterized CNN model depicted in Figure~\ref{fig:relation_module}. Like euclidean distance metric, it takes two input instances and estimates the distance between them. For example, let $f_{\varphi}(\mathbf{x}_i)$ and $f_{\varphi}(\mathbf{x}_j)$ be the feature embeddings of sample $x_i$ and $x_j$, the distance between them can be computed by Equation~\ref{eq:relation} as follows:
\begin{equation}
   d_{i,j} = g_{\phi}(\mathcal{O}(f_{\varphi}(\mathbf{x}_i),f_{\varphi}(\mathbf{x}_j))), ~~~\forall~i,j
\label{eq:relation}
\end{equation}
where $\mathcal{O}(\cdot, \cdot)$ denotes the preprocessing before pairwise comparison. We use the simple difference operation as defined in Equation~\ref{eq:rn_preprocessing} for preprocessing:
\begin{equation}
   \mathcal{O}(f_{\varphi}(\mathbf{x}_i),f_{\varphi}(\mathbf{x}_j)) = f_{\varphi}(\mathbf{x}_i) - f_{\varphi}(\mathbf{x}_j), ~~~\forall~i,j
\label{eq:rn_preprocessing}
\end{equation}
As shown in Figure~\ref{fig:framework}, the relation module is utilized as a shared submodule in the information aggregation procedure and the metric-based classifier for relation reasoning and classification respectively. That is:
\begin{equation}
\begin{aligned}
   & D_{\mathcal{C}} = g_{\phi}\circ\mathcal{O} \\
   & D_{\mathcal{G}} = g_{\phi}\circ\mathcal{O}
\end{aligned}
\label{eq:metric_assign}
\end{equation}
We will show in our experiments that the use of relation module in the information propagation and metric-based classifier boosts performance in comparison to that with plain euclidean distance metric.

\section{Experiments}
In this section, we evaluated and compared our MRN with state-of-the-art approaches on two few-shot learning benchmark datasets, \textit{i.e.} \textit{mini}Imagenet~\cite{snell2017prototypical} and \textit{tiered}Imagenet~\cite{ren2018meta}. 

\subsection{Datasets}
\noindent\textbf{\textit{mini}Imagenet} The \textit{mini}Imagenet dataset, originally introduced by Vinyals \etal~\cite{vinyals2016matching}, is derived from the larger ILSVRC-12 dataset~\cite{russakovsky2015imagenet}. It consists of 60,000 color images of size $84 \times 84$ that are divided into 100 categories with 600 examples each. We follow \cite{ravi2016optimization,snell2017prototypical} to split the dataset into 64 base categories for training, 16 novel categories for validation, and 20 novel categories for testing, respectively. In few-shot learning , the target model is trained on images from the 64 training categories and evaluated on the 20 novel categories. The validation set is used for monitoring generalization performance only.

\noindent\textbf{\textit{tiered}Imagenet} The \textit{tiered}Imagenet~\cite{ren2018meta} is another subset of the ILSVRC-12 dataset~\cite{russakovsky2015imagenet}, but it has a much larger class cardinality (608 classes) than that of \textit{mini}Imagenet. The dataset contains more than 700,000 images in total. The average number of examples in each class is more than 1200. Each of its 608 classes is derived from 34 high-level categories in Imagenet. The 34 high-level categories are further divided into 20 training (351 classes), 6 validation (97 classes) and 8 test (160 classes) categories. This high-level split strategy provides a more challenging and realistic few-shot setting where the training classes are distinct from the test classes semantically. 

\subsection{Implementation Details}
\noindent\textbf{Network architecture} For a fair comparison with existing methods, we took the widely-used four-layer convolutional network \cite{snell2017prototypical,finn2017model} Conv-4-64 as the basic feature extractor in our model. The backbone contains 4 convolutional blocks. Each block has a 64-filter convolutional layer with kernel size 3, a batch normalization layer, a relu activation layer, and a $2 \times 2$ max pooling layer. We utilize the relation module depicted in Figure~\ref{fig:relation_module} (as mentioned in Section~\ref{sec:relation_module}) for relation reasoning. The number of filters in each block of the relation module is set to 64 (\textit{i.e.}, $H=64$) in order to work with the feature extractor. 

\noindent\textbf{Episodic training} We adopt the episodic meta-training strategy proposed in \cite{vinyals2016matching} for rapid learning in our $K$-shot $C$-way experiments. In each training episode, besides the $K \times C$ support images, the \textbf{1-shot 5-way} episode contains 15 query images per each of the $C$ sampled categories. The number in the \textbf{5-shot 5-way} episode is 10 accordingly. It means one episode in total has $15 \times 5+1 \times 5=80$ training images for 1-shot 5-way experiments. All images are resized to $84 \times 84$. In the training phase, we also do basic data augmentations like random cropping and horizontal flipping to increase intra-class variations. The model is optimized with Adam~\cite{finn2017model} optimizer end-to-end from scratch. The learning rate is initially set to 0.001, and weight decay is set to $1 \times 10^{-6}$. For experiments on \textit{mini}Imagenet, we trained the model over 200,000 randomly sampled episodes and reduced the learning rate by half for every 50,000 episodes. As for experiments on \textit{tiered}Imagenet, we trained the model over 500,000 randomly sampled training episodes and reduced the learning rate by half for every 100,000 episodes. All our experiments are implemented in Pytorch with a GeForce GTX 1080 Ti Nvidia GPU card.

\noindent\textbf{Hyperparameters} We estimated the hyperparameters by cross validation on \textit{mini}Imagenet. In all our experiments, we set the neighborhood discovery hyperparameter $k=20$, information aggregation hyperparameter/search depth $d=1$, and information preservation hyperparameter $\lambda=0.2$. Namely, we selected the 20-nearest neighbors of an example as its neighborhood, and only propagated information from its 1-order neighbors for embedding enhancement even through information aggregation from neighborhood with arbitrary depth is possible.
After the aggregation procedure in Equation~\ref{eq:weighted_aggregation}, an embedding consists of 20 percents of old information and 80 percents of new information aggregated from its neighborhood. 

\noindent\textbf{Evaluation} We batched 15/10 query images per category in each episode for evaluation in 1-shot/5-shot 5-way classification. In all settings, we conducted few-shot classification on 1000 randomly sampled episodes from the test set and reported the mean accuracy together with the $95\%$ confidence interval.

\subsection{Baseline Methods}
We mainly compare with state-of-the-art methods that concentrate on representation learning or metric learnning to demonstrate the efficiency of our proposed MRN. Since we aim to extend Relation Network (RN)~\cite{sung2018learning} to exploit pairwise relationships for embedding enhancement, RN and its variant PARN~\cite{ziyang2019parn} would serve as the direct baselines. We also take TPN~\cite{liu2018learning}, MNE~\cite{li2019memory}, FEAT~\cite{ye2018learning}, GNN~\cite{garcia2017few} and MM-Net~\cite{qi2018mmnet} as baselines because our MRN shares some commonalities with them. To put it in a nutshell, we all propose to exploit the working context, the labelled instances or the unlabeled instances or them both, to boost few-shot learning in either a transductive manner or a non-transductive manner. Among them, the transductive methods, \ie TPN~\cite{liu2018learning}, MNE~\cite{li2019memory} and FEAT~\cite{ye2018learning}, taking advantage of both labelled and unlabeled instances, provide quite strong baselines.

We additionally provide MRN-Zero and MRN-Euclid as two baselines. MRN-Zero represents the variant that discards the working memory to avoid information propagation. With no memory, it loses the ability to aggregate information from other instances and inevitably falls back to the plain relation work. But MRN-Zero is slightly different from RN in that its relation module fuses two embeddings via differentiation preprocessing before pairwise comparison, rather than the concatenation in RN. MRN-Euclid shares a similar structure with our MRN except that it utilizes euclidean distance metric for relation estimation during the information aggregation procedure instead of the learnt one, which decouples the information propagation from distance metric learning. By introducing MRN-Euclid, we aim to demonstrate the superiority of the proposed MRN that has a compact structure and tightly coupled workflow.

\subsection{Main Results}
\label{sec:compare_with_sota}

\begin{table*}[]
   \begin{center}
   \resizebox{\linewidth}{!}{
   \begin{tabular}{lcccccccc}
      \hline \noalign{\smallskip}
      \multicolumn{5}{c}{} & \multicolumn{2}{c}{5-way Acc ($\%$, \textit{mini}Imagenet)} & \multicolumn{2}{c}{5-way Acc ($\%$, \textit{tiered}Imagenet)} \\
      Model & Ref & Backbone & Trans & Finetune  & 1-shot        & 5-shot   & 1-shot        & 5-shot     \\
      \hline \noalign{\smallskip}
      $\textbf{Meta-learning approaches}$ \\
      $\text{MAML}$~\cite{finn2017model} & ICML' 17 & Conv-4-64 & N & Y & $48.70 \pm 1.84$ & $63.11 \pm 0.92$ & $51.67 \pm 1.81$ & $70.30 \pm 1.75$ \\
      \hline \noalign{\smallskip}
      $\textbf{Metric-learning approaches}$ \\
      $\text{ProtoNet}$~\cite{snell2017prototypical} & NeurIPS' 17 & Conv-4-64 & N & N & $46.14 \pm 0.77$  & $65.77 \pm 0.70$ &  $48.58 \pm 0.87$  & $69.57 \pm 0.75$ \\
      $\text{ProtoNet(C+)}$~\cite{snell2017prototypical} &  NeurIPS' 17 & Conv-4-64 & N & N & $49.42 \pm 0.78$  & $68.20 \pm 0.66$ & - & - \\
      $\text{GNN}$~\cite{garcia2017few} & ICLR' 18 & Conv-4-256 & N & N & $50.33 \pm 0.36$ &  $66.41 \pm 0.63$ & - & -\\
      $\text{RN}$~\cite{sung2018learning} & CVPR' 18 & Conv-4-64 & N & N & $51.38 \pm 0.82$ & $67.07 \pm 0.69$ & $54.48\pm 0.93$ & $71.31\pm0.78$\\
      $\text{MM-Net}$~\cite{qi2018mmnet} & CVPR' 18 & Conv-4-64 & N & N & $53.37 \pm 0.48$ & $66.97 \pm 0.35$ & - & - \\
      $\text{DN4}$~\cite{li2019revisiting} & CVPR' 19 & Conv-4-64 & N & N & $51.24 \pm 0.74 $ & $71.02 \pm 0.64$  & - & - \\
      $\text{PARN}$~\cite{ziyang2019parn} & ICCV' 19 & Conv-4-64 & N & N & $55.22 \pm 0.84$   & $71.55 \pm 0.66$ & - & - \\
      $\text{TPN}$~\cite{liu2018learning}$^\dagger$ & ICLR' 19 & Conv-4-64 & Y & N & $53.75 \pm 0.86$ & $69.43 \pm 0.67$ & $57.53 \pm 0.96$ & $72.85 \pm 0.74$ \\
      $\text{TPN(K+)}$~\cite{liu2018learning}$^\dagger$ & ICLR' 19 & Conv-4-64 & Y & N & $55.51 \pm 0.86$ & $69.86 \pm 0.65$ & $59.91 \pm 0.94$ & $73.30 \pm 0.75$ \\
      $\text{MNE}$~\cite{li2019memory}$^\dagger$ & ICCV' 19 & Conv-4-64 & Y & N & $\mathbf{60.20} \pm 0.23$ & $72.16 \pm 0.17$ &  $60.04 \pm 0.28$ & $73.63 \pm 0.21$ \\
      $\text{FEAT}$~\cite{ye2018learning} & CVPR' 20 & Conv-4-64 & N & N & $55.15 \pm - $ & $71.61 \pm - $ & - & - \\
      $\text{FEAT}$~\cite{ye2018learning}$^{\dagger}$  & CVPR' 20 & Conv-4-64 & Y & N & $57.04 \pm 0.20$ & $\mathbf{72.89} \pm 0.16$ & - & - \\ 
      \hline \noalign{\smallskip}
      $\textbf{Hallucination based approaches}$ \\
      $\text{ActivationNet}$~\cite{qiao2018few} & CVPR' 18 & Conv-4-64 & N & N & $54.53 \pm 0.40$ & $67.87 \pm 0.20$ & - & - \\

      \hline \noalign{\smallskip}  
      $\text{MRN-Zero(Ours)}$ & & Conv-4-64 & N & N & $51.62 \pm 0.63 $ & $67.06 \pm 0.55 $ & $55.22 \pm 0.70$ & $73.37 \pm 0.62$ \\
      $\text{MRN-Euclid(Ours)}^{\dagger}$ & & Conv-4-64 & Y & N & $53.35 \pm 0.69 $ & $65.82 \pm 0.53 $ & $60.20 \pm 0.82$ & $71.94 \pm 0.65$ \\
      $\text{MRN(Ours)}^{\dagger}$ & & Conv-4-64 & Y & N &  $57.83 \pm 0.69 $ & $71.13 \pm 0.50 $ & $\mathbf{62.65} \pm 0.84$ & $ \mathbf{74.20} \pm 0.64 $ \\
      \hline
   \end{tabular}
   }
   \end{center}
   \caption{5-way classification accuracies of the proposed MRN and state-of-the-art methods on \textit{mini}Imagenet and \textit{tiered}Imagenet, with $95\%$ confidence intervals. Approaches fall into differented groups are separated, and top results are highlighted. $\text{C+}$: trained with higher way, $\text{K+}$: trained with higher shot, $\dagger$: transductive method, $\text{-}$: not reported.}
   \vspace{-1em}
   \label{table:exp_results}
\end{table*}

We validate the effectiveness of the proposed MRN on standard \textit{mini}Imagenet and \textit{tiered}Imagenet datasets, and report the main results in Table~\ref{table:exp_results}. 

\noindent\textbf{Comparison to previous state-of-the-arts} As shown in Table~\ref{table:exp_results}, MRN achieves $57.83\%/71.13\%$ classification accuracy for 1-shot/5-shot 5-way tasks on \textit{mini}Imagenet, and achieves $62.65\%/74.20\%$ classification accuracy for 1-shot/5-shot 5-way tasks on \textit{tiered}Imagenet. It outperforms non-transductive methods, such as ProtoNet, RN, GNN, DN4 and PARN, by a large margin on both datasets, which comfirms that exploring the working context by explicitly aggregating visual information from unlabeled instances benefits few-shot learning a lot. Thus the effectiveness of our MRN is proved. When compared to transductive methods, MRN largely outperforms TPN and its high-shot variant in both 1-shot and 5-shot learning on two datasets, and compares favorably with FEAT in 1-shot setting. MNE provides a strong baseline that is better than other previous work. In comparison, the performance of MRN is inferior to that of MNE on \textit{mini}Imagenet. But, our MRN gets more than $2\%/0.5\%$ improvement over MNE in 1-shot/5-shot setting on \textit{tiered}Imagenet and achieves state-of-the-art performance. It should be noted that the MRN tends to achieve higher improvements or better results on \textit{tiered}Imagenet than on \textit{mini}Imagenet. We conjecture this is primarily due to there existing more categories and more training examples in the training split of \textit{tiered}Imagenet dataset which in turn presents richer intra-class variations that facilitates model training. We also observe that MRN consistently makes more gains for 1-shot learning in comparison to its 5-shot counterpart, which indicates that the proposed embedding enhancement in Section~\ref{sec:embedding_enhancement_within_memory} is more helpful when training data is extremely scarce.

\noindent\textbf{Comparison among MRN variants} Among the three MRN variants, MEN-Zero refuses to propagate information from other instances for embedding enhancement and consequently achieves inferior performance when compared to previous transductive methods. This fits in with expectation because transductive methods consistently outperform non-transductive methods in few-shot learning. When compared to RN, we can observe that it gets competitive results on \textit{mini}Imagenet and better results on \textit{tiered}Imagenet, which demonstrates the effectiveness of differentiation preprocessing used in our relation module as described in Section~\ref{sec:relation_module}. MRN-Euclid outperforms MRN-Zero by a large margin on \textit{mini}Imagenet and \textit{tiered}Imagenet for 1-shot learning, but surprisingly falls behind MRN-Zero on both datasets by around $1.3\%$ for 5-shot learning. We conclude this is mostly due to its decoupled workflow where euclidean distance is adopted for relation estimation during embedding enhancement and decouples the embedding enhancement from distance metric learning. In contrast, MRN consistently gets improvements, demonstrating the superiority of our compact model.

\begin{figure}
   \begin{center}
      \includegraphics[width=\linewidth]{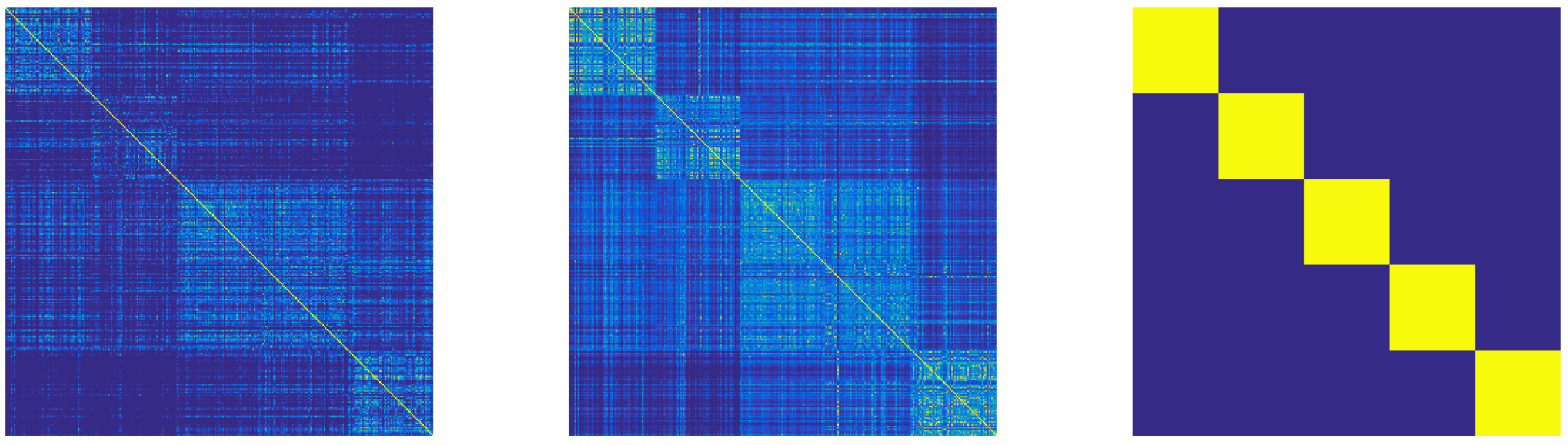}
   \end{center}
   \vspace{-1em}
   \caption{ Similarity matrices under 5-shot 5-way setting on \textit{mini}Imagenet. From left to right: RN, our MRN and the ground truth. A brighter color indicates a higher estimated similarity score.}
   \label{fig:similarity_matrix}
\end{figure}

\begin{figure*}
   \begin{center}
      \includegraphics[width=\linewidth]{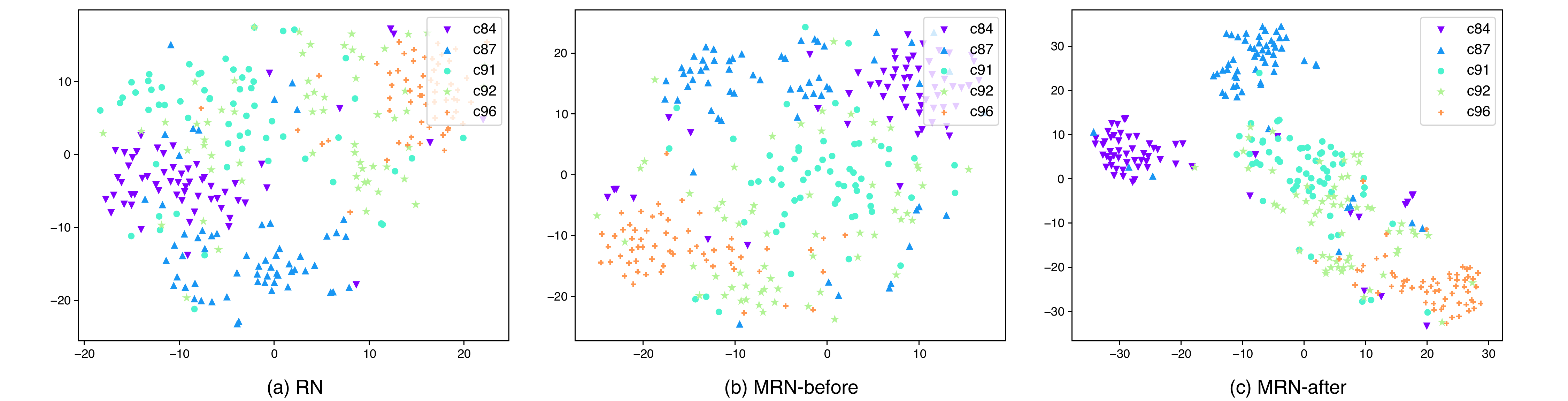}
   \end{center}
   \vspace{-1em}
   \caption{ t-SNE visualization under 1-shot 5-way setting on \textit{mini}Imagenet. From left to right: (a) RN, (b) MRN before propagation and (c) MRN after propagation.}
   \label{fig:feat_vis}
\end{figure*}

\noindent\textbf{Visualization}
We visualize the similarity matrices learned by RN and our MRN under 5-shot 5-way setting on \textit{mini}Imagenet. In detail, we randomly sampled a 5-shot 5-way task from the test split. For each category in the task, we select 10 query instances as in the training phase. Then all instances are fed into the target models for generating the matrices. Given the feature representations extracted by the backbone, RN computes its similarity matrix based on the pairwise similarities estimated by the learnt distance metric. Our MRN performs weighted information propagation as described in Section~\ref{sec:embedding_enhancement_within_memory} to enhance representations right before it estimates the similarities. We visualize the $75 \times 75$ matrices in Figure~\ref{fig:similarity_matrix}. It can be seen that MRN generates a matrix that is much closer to the ground truth. We also sampled a 1-shot 5-way task with 59 query images per each category, extracted the features and visualized them using t-SNE in Figure~\ref{fig:feat_vis}. The first information we can get from the figure is that the backbone or the embedding function can effectively learn to map similar images to nearby points in the embedding space (see Fighre~\ref{fig:feat_vis}(a)). From Figure~\ref{fig:feat_vis}(c), we can observe that the propagation procedure dramatically decreases the intra-class variations and increases the inter-class variations. The embeddings after explicitly aggregating information from other instances in the working context are shown to be more discriminative.

\subsection{Ablation Study}
\label{sec:ablation_study}
To understand MRN better, we carried out several controlled experiments to examine how each part affects the final performance.


\begin{figure*}[t]
   \begin{center}
      \includegraphics[width=\linewidth]{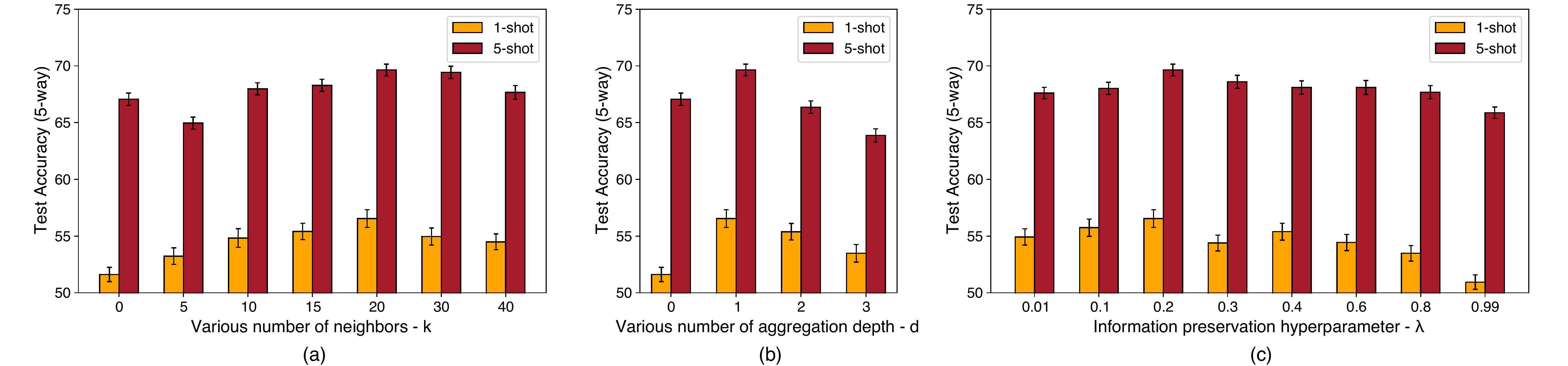}
   \end{center}
   \vspace{-1em}
   \caption{The changes of 5-way classification accuracy on \textit{mini}Imagenet as (a) number of neighbors increases, or (b) the number of search depth increases, or (c) the amount of preserved information increases in the information aggregation procedure.}
   \vspace{-0.5em}
   \label{fig:ablation_study}
\end{figure*}

\noindent\textbf{Sensitivities to hyperparameters} The number of neighbors, \textit{i.e.} the neighborhood discovery hyperparameter $k$, is vital for information propagation. In Figure~\ref{fig:ablation_study}(a), we investigate the setting of $k$. It can be observed that MRN with $k>0$ always surpass MRN-Zero (the baseline model or the plain relation network with $k=0$). We conclude that embedding enhancement by aggregating information from the neighborhood helps. But, we also notice that the gain drops when too few ($k < 10$) or too many ($k>30$) neighbors are exploited. This is mainly because a tiny neighborhood can only provide limited amount of information and a huge neighborhood imposes irrelevant information which makes it difficult for our MRN to further boost performance.

In Figure~\ref{fig:ablation_study}(b), we compare different values of $d$, the information aggregation hyperparameter used in Equation~\ref{eq:iterative_aggregation} that controls how deep into the relation graph we go. By increasing $d$, we iteratively collect more information from less relevant instances. From the results, we can see that the performance tends to decrease as the search depth $d$ increases. For 5-shot classification, the performance of MRN with $d=3$ is even worse than that of the plain relation network MRN-Zero with $d=0$, though it still achieves superior performance for 1-shot classification.


\noindent\textbf{Impact of information preservation} In Equation~\ref{eq:weighted_aggregation}, we set the hyperparameter $\lambda$ for information preservation. A smaller $\lambda$ means more information aggregated from the neighborhood, and correspondingly a bigger $\lambda$ means more information from the instance itself. To see how the performance changes along with $\lambda$, we conducted sensitivity experiments as shown in Figure~\ref{fig:ablation_study}(c). As can be seen from the results, the performance drops slightly in 5-shot setting and drastically in 1-shot setting when less and less information is aggregated from the neighborhood. However, our MRN is able to perform neck to neck with MRN-Zero even when $\lambda = 0.99$ in which situation nearly no information is aggregated. Quite unexpectedly, we find that the MRN outperforms MRN-Zero when only one percent of the original information is preserved, with $99$ percents of the information aggregated from the neighbors. This indicates the fact that the proposed MRN is effective enough to discriminate relevant examples from irrelevant ones and borrow information from the relevant examples to form more discriminative embeddings.

\begin{table}[]
   \begin{center}
      \resizebox{\linewidth}{!}{
      \begin{tabular}{lcccc}
         \hline \noalign{\smallskip}
         \multicolumn{3}{c}{} & \multicolumn{2}{c}{5-way Acc ($\%$)} \\
         Model & Mem-Aug & Info-Agg  & 1-shot        & 5-shot \\
         \hline \noalign{\smallskip}
         $\text{MRN-Zero}$ & No & No & $51.62 \pm 0.63$ & $67.06 \pm 0.55$ \\
         \hline \noalign{\smallskip}
         $\text{MRN-mean}^\dagger$ & Yes & mean & $52.33 \pm 0.75$ & $63.05 \pm 0.52$ \\
         $\text{MRN-max}^\dagger$ & Yes & max & $51.28 \pm 0.71$ & $63.71 \pm 0.53$\\
         $\text{MRN(Ours)}^\dagger$ & Yes & weighted & $57.83 \pm 0.69 $ & $71.13 \pm 0.50 $ \\
         \hline
      \end{tabular}}
   \end{center}
   \caption{5-way classification results on \textit{mini}Imagenet with different information aggregation strategies.}
   \vspace{-1em}
   \label{table:exp_as_aggregation}
\end{table}
\noindent\textbf{Effectiveness of weighted embedding aggregation} General aggregation strategies like mean-pooling and max-pooling can be used in the information aggregation procedure. In Table~\ref{table:exp_as_aggregation}, we compared the proposed weighted embedding aggregation with mean-pooling and max-pooling feature aggregation methods. MRN-mean is the model that employs mean pooling aggregation strategy during the information aggregation procedure. MRN-max denotes the one that uses max pooling aggregation strategy. The performance drops drastically for both MRN-mean and MRN-max. For MRN-mean, aggregating information equally from all neighbors hurts performance because the aggregation imposes too much irrelevant information. For MRN-max, the performance drop results from information loss when only the max values are selected and merged into the target representations.

\section{Conclusion}
In this work, we tackled few-shot learning problems from the perspective of metric learning. Different from previous work that adopt a predefined metric, such as euclidean distance or cosine distance, in the metric-based classifier for similarity measurement, we learn a generic distance metric to compare images for the same purpose. We further propose to propagate information from relevant instances in working context based on their visual affinities for embedding enhancement. We introduce the episodic memory to serve as the working context in which pairwise relationships between each two instances are estimated with the learnt metric and the representation of an instance is enhanced by attentively aggregating information from its neighborhood. We empirically demonstrate that the learnt distance metric suits few-shot learning better than the predefined ones and the enhanced embeddings are more discriminative which in turn boosts performance. The proposed MRN with this tightly coupled workflow achieved state-of-the-art results on the benchmark datasets. For future work, we plan to extend MRN to deeper models. We also notice that relation propagation is a promissing direction and will study it as a supplement to our visual information propagation in the near future.


\bibliographystyle{ACM-Reference-Format}


\end{document}